\documentclass[letterpaper, 10 pt, conference]{ieeeconf}  

\IEEEoverridecommandlockouts                              

\usepackage{comment}
\usepackage{amsmath,amssymb} 
\usepackage{booktabs}
\usepackage{multirow}
\usepackage{multicol}
\usepackage{adjustbox}
\usepackage{newtxtext,newtxmath}
\usepackage{graphicx}
\usepackage{booktabs,multirow,tabularx,array}
\usepackage{subcaption}
\usepackage{cleveref}
\usepackage{stfloats}

\newcolumntype{L}[1]{>{\raggedright\arraybackslash}p{#1}} 
\newcolumntype{Y}{>{\centering\arraybackslash}X}  

\title{\LARGE \bf
Introspection in Learned Semantic Scene Graph Localisation 
}

\author{Manshika Charvi Bissessur, Efimia Panagiotaki, Daniele De Martini \\ 
Mobile Robotics Group (MRG), University of Oxford, UK
\thanks{The project was supported by the EPSRC Programme Grant ``From Sensing to Collaboration'' (EP/V000748/1). Manshika Charvi Bissessur was supported by the State of Mauritius Science Scholarship, and Efimia Panagiotaki by the Oxford-DeepMind Scholarship. Corresponding author: Manshika Charvi Bissessur,  \texttt{manshika.bissessur@reuben.ox.ac.uk}
}}

\begin{document}

\maketitle
\thispagestyle{empty}
\pagestyle{empty}

\begin{abstract}
This work investigates how semantics influence localisation performance and robustness in a learned self-supervised, contrastive semantic localisation framework.
After training a localisation network on both original and perturbed maps, we conduct a thorough post-hoc introspection analysis to probe whether the model filters environmental noise and prioritises distinctive landmarks over routine clutter.
We validate various interpretability methods and present a comparative reliability analysis. Integrated gradients and Attention Weights consistently emerge as the most reliable probes of learned behaviour. A semantic class ablation further reveals an implicit weighting in which frequent objects are often down-weighted. Overall, the results indicate that the model learns noise-robust, semantically salient relations about place definition, thereby enabling explainable registration under challenging visual and structural variations.
\end{abstract}

\section{Introduction}
Conventional localisation methods that rely on extracting geometric features from input sensor data, are often susceptible to perceptual variations. Factors such as changes in lighting and weather conditions, or the presence of dynamic objects can significantly alter the appearance of traffic scenes, making these methods less reliable.
In contrast, semantic information offers greater robustness, as it is inherently invariant to variations in environmental conditions. For instance, a desk remains identifiable as a desk regardless of whether it is observed in daylight or darkness, making it a stable and reliable landmark for localisation.
Notably, this is consistent with how humans navigate an environment, mainly relying on high-level semantic cues -- such as rooms, furniture, and structural layouts -- rather than low-level geometric attributes or ``pixel-level'' appearance features.

Reflecting this, recent works in place recognition and global localisation increasingly leverage higher-level visual features that correspond more closely to the semantic structure of the environment \cite{Liu,yoshida2023activesemanticlocalizationgraph,panagiotaki2023semgatexplainablesemanticpose,pramatarov2022boxgraph,pramatarov2024s}.
Beyond leveraging semantics, humans naturally filter out environmental ``noise'' to localise themselves, instinctively distinguishing scene elements that concretely define a place. While objects offer limited localisation value, persistent structures provide stable and reliable reference points. 

Prior research has demonstrated that semantic elements can be effectively exploited for localisation while also investigating their role in localisation performance and decision making \cite{panagiotaki2023semgatexplainablesemanticpose, pana2}. Building on this foundation, we extend the scope of the analysis by systematically examining model introspection. Specifically, we propose a semantics-driven framework in which a model performs place registration from high-level semantic layers, followed by thorough post-hoc introspection that quantifies the influence of each object class.
This class-level importance analysis enables us to verify that the model prioritises stable, meaningful features over transient clutter, mirroring the filtering strategies employed by humans.
Because these explanations are generated after inference, they provide an independent and interpretable audit trail, which is crucial for diagnosing failure modes early and building trust in the robot's decisions, especially in safety-critical and dynamic environments.

In this work, we address a critical need in current robotics research, advancing towards localisation systems that combine robustness and efficiency with transparency and interpretability.
Our key contributions are as follows:
\begin{enumerate}
  \item   
   A perturbation-based class-importance analysis, assessing both place registration performance degradation and attribution shifts;
 
  \item   
    A rigorous introspection analysis, along with fidelity tests, demonstrating that Integrated Gradients and Attention Weights provide the most faithful object-importance attribution signals.
\end{enumerate}

\section{Related Work}
\subsection{Semantic Localisation}
A common strategy for semantic localisation is to augment individual features \cite{Kobyshev} or bag-of-words (BoW) representations \cite{Arand} with semantic information as a post-processing step.
Going further, \cite{schönberger2018semanticvisuallocalization} learns a single descriptor that fuses semantics and geometry.
Hybrid approaches such as SemSegMap \cite{Cremariuc} leverage both cues and generally achieve better localisation than geometry alone.

There are also settings where only semantic information is available for the query \cite{naricp}. Text2Loc \cite{Xia_2024_CVPR} addresses this by aligning rich textual embeddings extracted with a frozen T5 model with 3D submap embeddings derived from point clouds using PointNet++, trained via a cross-modal contrastive objective.

\subsection{Importance of Different Semantic Classes}
Neural networks excel in a wide range of tasks, yet their distributed feature representations remain semantically opaque. Since the features leveraged by these models often lack a direct correspondence to human-interpretable concepts, explaining their decision-making processes can be challenging. 
In the domain of geolocalisation, \cite{Theiner_2022_WACV} introduces a concept-influence metric that estimates the contribution of semantic visual concepts (e.g., ``building'', ``mountain'') to a model's prediction -- rather than specific pixels or regions. Using Integrated Gradients with SmoothGrad to reduce noise, they find that localised concepts (e.g., tower, bridge) are highly informative. In contrast, generic concepts (e.g., grass, car) contribute little due to their ubiquity.

We extend this line of research to introspective localisation on scene graphs. The closest prior work is SEM-GAT \cite{panagiotaki2023semgatexplainablesemanticpose}, an attention-based GNN method that integrates semantic and geometric cues to identify reliable correspondences for point cloud registration also introducing an explainability component by analysing the model's attention weights \cite{pana2}. 
Our work focuses on indoor environments and investigates analogous semantic–structural relationships on scene graphs. We conduct a more comprehensive introspection analysis that goes beyond attention weights, proposing and evaluating a range of post-hoc explainability methods to identify reliable objects that can serve as strong landmarks for interpretable and accurate localisation.

\subsection{Model Explainability for Graphs}
\label{sec:explainlit}

Graph neural networks (GNNs) integrate rich node and edge features with graph topology, yielding strong performance but often lacking transparency in their decision-making processes.
We focus on post-hoc explanations, leaving the trained model unchanged, and follow the taxonomy of~\cite{Yuan}, which categorises methods into: (i) gradient/feature-based, estimating importance from input gradients or feature magnitudes; (ii) perturbation-based, measuring prediction changes when features, nodes, or edges are modified or removed; (iii) decomposition-based, tracing contributions backward through the network; and (iv) surrogate-based, fitting simple, interpretable models (e.g.,decision trees) to approximate the behaviour of GNNs.
Within this space, \textit{Saliency} \cite{simonyan2014deepinsideconvolutionalnetworks}, \textit{Integrated Gradients} \cite{sundararajan2017axiomaticattributiondeepnetworks}, and \textit{Shapley Value Sampling} \cite{trumbelj2010AnEE} are employed.
Attention weights are also considered as model-intrinsic signals in architectures such as GATs \cite{veličković2018graphattentionnetworks}. While early work cautioned that attention alone may be insufficient or misleading as an explanation~\cite{wen2022revisitingattentionweightsexplanations}, subsequent studies argue it can be informative when validated~\cite{wiegreffe2019attentionexplanation}. In this vein, \cite{pana2, fan2021gcnseattentionexplainabilitynode} recommend verifying that changes in attention correlate with performance before treating attention as an explanation.

\section{Overview}

\begin{figure}[!t]
  \centering
  \includegraphics[width=0.9\linewidth]{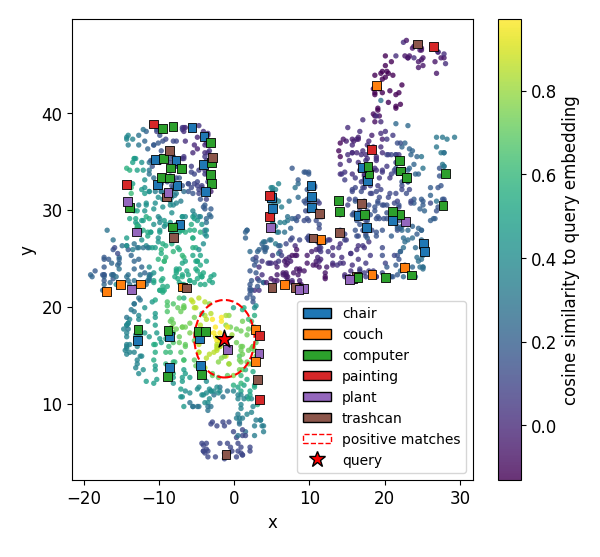}
  \vspace{-15pt}
  \caption{Floor-plan of the \textit{office} scene. Each L2 and L3 nodes are plotted in its metric coordinate. L3 nodes are coloured by cosine similarity to a query.}
  \label{fig:floorplan}
      \vspace{-15pt}
\end{figure}

In this section, we formulate the problem addressed and describe the scene graphs and dataset used. A 3D scene graph models an environment as a hierarchy of spatial concepts (nodes) and their relations (edges), providing a compact abstraction that captures both semantic cues and spatial organisation.

Our work leverages the hierarchical scene graphs proposed in Hydra \cite{hughes2022hydrarealtimespatialperception} which employs five layers of hierarchy: (L1) Metric-Semantic Mesh, a dense 3D mesh with per vertex geometry and semantic labels; (L2) Objects \& Agents, entities (including the robot) with spatial and semantic attributes, each linked to the mesh elements composing them; (L3) Places, free space nodes connected by traversability edges into a topological map and linked to nearby L2 nodes; (L4) Rooms, nodes with location and bounding boxes connected to their constituent places; (L5) Buildings, nodes connected to all constituent rooms.
Our experiments use only the L2-L3 layers to stress the model's semantic reliance.

The uHumans2 dataset \cite{Rosinol} provides photorealistic, synthetic indoor environments with complete metric–semantic annotations, making it a standard benchmark for spatial-perception pipelines such as Hydra, which produces a scene graph comprising about 1,300 place nodes and 110 object nodes on the \texttt{office} scene, depicted in \Cref{fig:floorplan}.

\begin{table}[h]
\centering
\caption{Breakdown of object instances in the uHumans2 office scene.}
\label{tab:uhumans2_objects}
\begin{tabular}{c c c}
\toprule
\shortstack{\textbf{Semantic}\\\textbf{Label}} & 
\shortstack{\textbf{Semantic}\\\textbf{Class}} & 
\shortstack{\textbf{Semantic}\\\textbf{Instances}} \\
\midrule

5 & Chair (CH)       & 28 \\
8 & Couch (CO)       & 11 \\
10 & Computer (CP)   & 35 \\
11 & Plant (PL)       & 13 \\
12 & Painting (PA)   & 8  \\
18 & Trash-can (TC)   & 15 \\
\bottomrule
\end{tabular}
\end{table}

Given the class distribution in \Cref{tab:uhumans2_objects} and its potential effect on localisation, we expect:
\begin{enumerate}
\item \textbf{Broad but moderate signals from frequent classes:} computers and chairs should offer wide coverage while contributing only moderate confidence to mitigate perceptual aliasing.
\item \textbf{High salience of distinctive landmarks:} rarer categories such as paintings and couches should act as strong, location-specific anchors for disambiguation.
\item \textbf{Down-weighting of mobile objects over time:} the relative importance of chairs and trash cans should decrease across traversals as positional variability is detected.
\end{enumerate}

These hypotheses are assessed via class-removal perturbations and post hoc attribution analyses.
\begin{table}[t]
  \centering
  \caption{Validation and test performance on uHumans2 (\texttt{office}).}
  \label{tab:mpnn_gat_results}
  \renewcommand{\arraystretch}{1.3}
  \resizebox{\columnwidth}{!}{%
  \addtolength{\tabcolsep}{-0.5em}
  \begin{tabular}{lccccccccccc}
    \toprule
    Model & Loss &
    \multicolumn{5}{c}{Validation} &
    \multicolumn{5}{c}{Testing} \\
    \cmidrule(lr){3-7}\cmidrule(lr){8-12}
     &  & PR-AUC & F1 & \multicolumn{3}{c}{Recall@N (\%)} & PR-AUC & F1 & \multicolumn{3}{c}{Recall@N (\%)} \\
    \cmidrule(lr){5-7}\cmidrule(lr){10-12}
     &  &  &  & 1 & 5 & 10 &  &  & 1 & 5 & 10 \\
    \midrule
    \multirow{3}{*}{Ours}
      & Contrastive & 0.667 & 0.649 & 8.92 & 39.82 & 63.62 & 0.721 & 0.680 & 17.22 & 63.12 & 85.12 \\
      & Triplet     & 0.524 & 0.554 & 8.86 & 37.21 & 59.53 & 0.562  & 0.571 & 17.59 & 59.82& 78.54 \\
      & InfoNCE     & 0.699 & 0.685 & 8.61 & 40.53 & 65.64 & 0.718 & 0.685 & 18.21 & 65.72 & 86.14 \\
    \addlinespace[0.6ex]
    BoW & -- & 0.247 & 0.338 & 5.32& 23.79& 38.37& 0.243& 0.360& 8.60& 32.68& 51.15  \\
    \bottomrule
  \end{tabular}%
  }
\end{table}

\section{Semantic Localisation}

\subsection{Problem Statement}
Let a large-scale 3D hierarchical scene graph $M = (\mathcal{P} \cup \mathcal{O}, \mathcal{E}^t \cup \mathcal{E}^v)$ be a subset of Hydra's \cite{hughes2022hydrarealtimespatialperception}, containing places $\mathcal{P} = \{p_i\}$ and objects $\mathcal{O} = \{o_i\}$, connected by traversability edges $\mathcal{E}^t = \{e^t_{i,k}\}$ between places $i$ and $k$, and visibility edges $\mathcal{E}^v = \{e^v_{i, j}\}$ between a place $i$ and an object $j$.
Each object and place are denoted by their Cartesian coordinates $(x, y)$ and each object also by its semantic class $c \in \mathcal{C}$, where $\mathcal{C}$ is the set of semantic classes.

Let \(T\) denote a query place.
The goal is to learn an encoder \(F\) that maps both $T$ and the reference places in $M$ into a metric space, such that the nearest neighbour of \(T\) in $M$ corresponds to its true location.
%
%
Additionally, we want to assign to each class \(c\in\mathcal{C}\) an
\emph{attribution} score that quantifies its contribution to localisation, both at the instance level and in expectation over the dataset.


\subsection{Semantic Localisation}
We employ a GNN backbone relying on message passing and attention, trained contrastively on two perturbed variants of the \texttt{office} dataset to form place–query pairs.


\subsubsection{GNN Backbone}
\label{sec:architecture}
We \textit{intentionally} represent each object node with only the semantic class associated with it and treat each place node as an aggregation point, discarding geometric information, to force our model to focus on the semantic cues.
Semantic labels are linearly projected into 64-dimensional hidden states and passed through an ELU activation.
Two subsequent message-passing layers aggregate neighbour information with a \texttt{sum} operator, each followed by $\tanh$ nonlinearity and batch normalisation.
Long-range context is captured with a \textsc{GATv2} convolution using three attention heads, and a final linear layer maps hidden states into 32-dimensional embeddings for localisation.

\subsubsection{Dataset Splits and Perturbations}
\label{sec:dataset}
Experiments use the \texttt{office} subset of uHumans2 with two repeats of the same trajectory. Place nodes are partitioned into 70-20-10 split for train, validation, and test respectively, with the test set unseen during training.
For data augmentation, object nodes are displaced according to class-specific mobility (e.g., small shifts for highly mobile classes such as trash cans and chairs), thereby generating additional query–target pairs and reducing overfitting to the original layout. Since Hydra often produces closely spaced places, a matching radius of \(4\,\mathrm{m}\) is used: all map nodes within this radius of the ground-truth location are treated as positives, allowing multiple positive matches per query.

\subsubsection{Metrics}
Evaluation is conducted from two complementary views.
\textit{Classification}: we report F1 and PR–AUC, as positives are sparse and the class distribution is highly imbalanced.
\textit{Retrieval}: we report Recall@\(\,N\,\), which reflects practical usage where a shortlist is inspected and quantifies the probability that a correct match appears among the top-\(N\) candidates. \Cref{tab:mpnn_gat_results} summarises our results on the uHumans2 dataset when applying different losses -- contrastive \cite{Hadsell2006DimensionalityRB} with margin of 1, triplet loss \cite{tripletweinberger09a} with hard mining, and InfoNCE ($T=0.7$) \cite{oord_representation_2018}.
Among these, InfoNCE yields the most balanced performance across PR-AUC, F1, and Recall@\(N\).
Moderate absolute scores are expected under a semantics-only regime on a small dataset, as discarding geometry removes precise 3D anchors, label aliasing occurs when visually distinct objects share the same class, and limited training diversity restricts generalisation. \textit{The semantics-only constraint is intentional}, allowing an isolated assessment of semantic cues.
For context, \cref{tab:mpnn_gat_results} compares the model with a bag-of-words (BoW) baseline that represents each location by per-class object counts; the learned encoder improves PR-AUC by a large margin and consistently increases Recall@N. 
Although absolute recall remains modest, the relative gains indicate a more discriminative semantic embedding for retrieval.

\begin{figure}[!t]
  \centering
  \subfloat[Ground truth.\label{fig:gt}]{\includegraphics[width=0.32\linewidth]{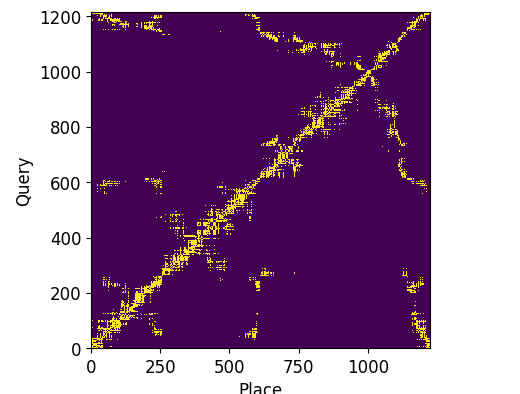}}\hfill
  \subfloat[BoW similarity.\label{fig:baselinesimilarity}]{\includegraphics[width=0.32\linewidth]{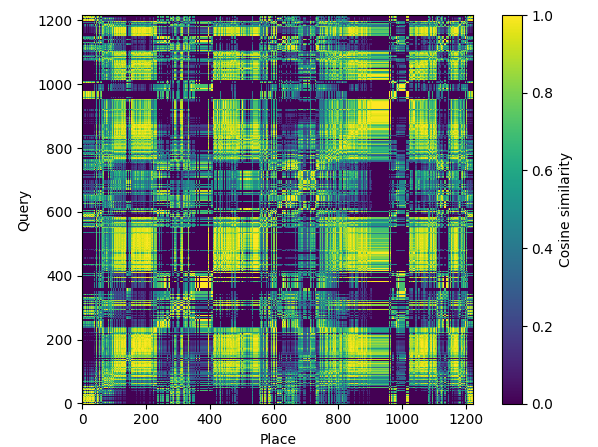}}\hfill
  \subfloat[Model similarity.\label{fig:modelsimilarity}]{\includegraphics[width=0.32\linewidth]{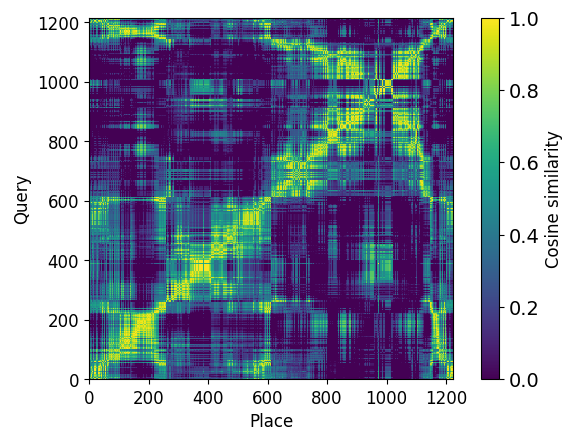}}
  \caption{Place--query matrices. Diagonal structure indicates correct matches; our model sharpens the diagonal and suppresses false positives.}
    \vspace{-15pt}

\end{figure}

Qualitative results support these findings. The similarity matrices in Figs.~\ref{fig:gt}--\ref{fig:modelsimilarity} reveal a clear diagonal of correct matches; the learned encoder sharpens this diagonal and suppresses off-diagonal responses, improving separation of non-matching pairs. Unlike the BoW baseline, which primarily responds to frequent classes and thus conflates semantically similar yet spatially distinct locations, the learned localiser adaptively reweights object configurations around each query, amplifying distinctive object–object relations (e.g., a specific couch–painting pairing) while down-weighting ubiquitous items (e.g., computers).
\Cref{tab:ablation} shows that removing the GAT block induces a marked performance drop that is not recovered by adding MPNN \cite{mpnn}. 

\begin{table}[t]
\centering
\caption{Ablation study: impact of model design choices on PR-AUC~\((\uparrow\)) and Recall@1~\((\uparrow\)). Best score in \textbf{bold}.}
\label{tab:ablation}
\renewcommand{\arraystretch}{1.05}
\resizebox{\columnwidth}{!}{%
\addtolength{\tabcolsep}{-0.5em}
\begin{tabular}{lccccc}
\toprule
Variant & \# MPNN & Hidden Dim & Heads & PR-AUC & Recall@1 \\
\midrule
Number of layers            & 1  & 64  & 3  & 0.34 $\pm$ 0.03 & 0.14 $\pm$ 0.05 \\
           & 3  & 64  & 3  & 0.54 $\pm$ 0.11                   & 0.15 $\pm$ 0.01 \\
Hidden dimension          & 2 & 32  & 3  & 0.63 $\pm$ 0.08                   & 0.16 $\pm$ 0.01 \\
         & 2 & 128 & 3  & \textbf{0.73 $\pm$ 0.03}          & \textbf{0.17 $\pm$ 0.01} \\
Attention heads             & 2 & 64  & 1  & 0.58 $\pm$ 0.05                   & 0.14 $\pm$ 0.01 \\
         & 2 & 64  & 2  & 0.65 $\pm$ 0.07                   & 0.14 $\pm$ 0.02 \\
No GAT block  & 2   & 64  & -- & 0.09 $\pm$ 0.01  & 0.02 $\pm$ 0.003 \\
  & 3   & 64  & -- & 0.61 $\pm$ 0.02                   & 0.16 $\pm$ 0.01 \\
  \midrule
\textbf{Our model} & \textbf{2} & \textbf{64} & \textbf{3} & 0.72 $\pm$ 0.02 & \textbf{0.17 $\pm$ 0.02} \\
\bottomrule
\end{tabular}%
}
\end{table}

\section{Semantic Class Attribution}

The role of object semantics in localisation is quantified through complementary analyses, including class ablation, post-hoc attribution, and fidelity analysis, with no retraining process.

\textbf{Class ablation} For each semantic class \(c\), all instances are removed, and we measure the change in performance as the difference in PR-AUC with and without the objects of class $c$, normalised by normalised by class frequency.

\textbf{Post-hoc attribution} Node-level importance is estimated using Saliency, Integrated Gradients, Shapley Value Sampling, and Attention Weights. Let \(p^{\text{pre}}\) and \(p^{\text{post}}_c\) denote the distributions of node-importance scores before and after perturbing class \(c\), respectively. The shift is quantified by the Jensen–Shannon divergence (JSD) between \(p^{\text{pre}}\) and \(p^{\text{post}}_c\), normalised by class frequency.
A larger JSD indicates a greater criticality of class \(c\) (i.e., its removal substantially shifts the attribution scores) and reveals how its absence reweights the remaining nodes.

\textbf{Fidelity} Fidelity of the explanation \cite{Pope}, i.e.whether highlighted nodes influence the output, is assessed via \textit{necessity} and \textit{sufficiency} tests, also called \textit{fidelity\(^{+}\)} and \textit{fidelity\(^{-}\)}, adapted to the contrastive objective, using changes in embedding similarity rather than predicted probabilities.

Let \(\mathbf z:\mathcal G \to \mathbb R^d\) denote the learned encoder that maps a place graph \(G\) to a \(d\)-dimensional embedding.
For a place graph \(P\) and its perturbed query \(Q\), let the baseline similarity $s_{\mathrm{full}}$ be the scalar product of $\mathbf{z}$ applied to $P$ and $Q$.
We can then define a budget $\rho$ to be the fraction of nodes of $Q$ we can keep, ranked by an explainer.
We can then recalculate the similarity between $\mathbf{z}$ applied to $P$ and $Q$ containing only the top nodes selected by $\rho$ or every node excet them.
We the two metrics with $s_{\mathrm{keep}}(\rho)$ and $s_{\mathrm{drop}}(\rho)$ respectively.


$s_{\mathrm{keep}}(\rho)$ and $s_{\mathrm{drop}}(\rho)$ can be used to calculate the fidelity\(^{+}\) (necessity) and fidelity\(^{-}\) (sufficiency) of $\mathbf{z}$ as:
\begin{equation}
\begin{aligned}
\Delta^{+}(\rho)\;=\;\bigl|\, s_{\mathrm{drop}}(\rho) - s_{\mathrm{full}} \,\bigr| \\
\Delta^{-}(\rho)\;=\;\bigl|\, s_{\mathrm{full}} - s_{\mathrm{keep}}(\rho) \,\bigr|
\end{aligned}
\end{equation}
Large \(\Delta^{+}(\rho)\) indicates that the removed top-ranked nodes are necessary, while small \(\Delta^{-}(\rho)\) indicates that the retained top-ranked nodes are sufficient.

\noindent Finally, at a fixed budget \(\rho_{\star}\) we aggregate necessity and sufficiency into a single \emph{characterisation score} using a weighted harmonic mean of \(\Delta^{+}\) and \(1-\Delta^{-}\):
\begin{equation}
  \mathrm{charact}(w_{+},w_{-};\rho_{\star})
  \;=\;
  \frac{w_{+} + w_{-}}
       {\displaystyle\frac{w_{+}}{\Delta^{+}(\rho_{\star})}
        \;+\;
        \frac{w_{-}}{1 - \Delta^{-}(\rho_{\star})}}
  \label{eq:char_score}
\end{equation}
with \(w_{+},w_{-}\in[0,1]\) and \(w_{+}+w_{-}=1\). Higher \(\mathrm{charact}\) indicates that the explainer simultaneously exhibits strong necessity and sufficiency at budget \(\rho_{\star}\).

\section{Results}

%
%
\begin{figure}[t]
    \includegraphics[scale = 1.0]{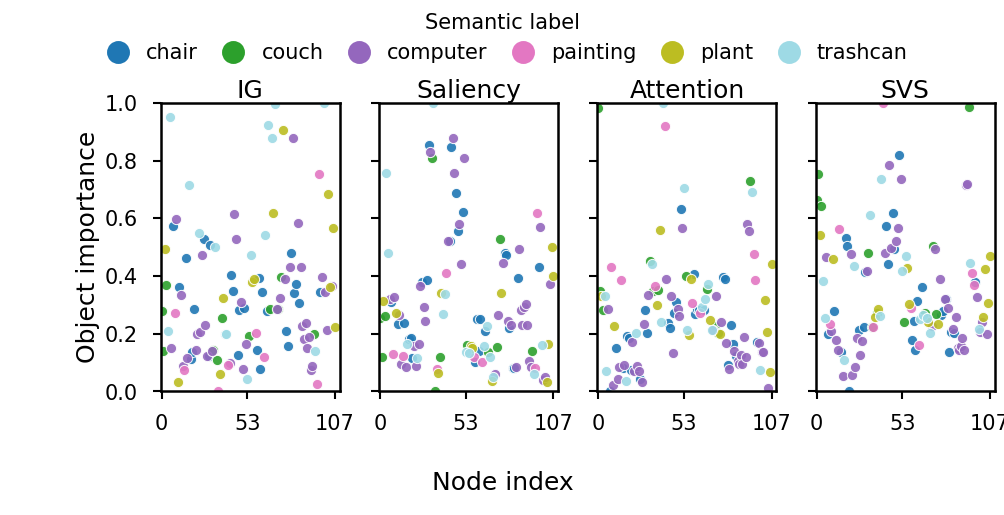}
    \centering
    \caption[Object node importance cd across places]{Importance of each object node averaged over all place embeddings in the office scene, as determined by four explainability methods: Saliency, Integrated Gradients, Attention Weights, and Shapley Value Sampling. All methods highlight a small subset of highly influential objects, but notable discrepancies in mid-range scores reveal method-specific differences in assessing node relevance.}

    \label{fig:objimp}
    \vspace{-15pt}
\end{figure}

\subsection{Class Ablation}
As shown in \cref{tab:all-rankings}, leave-one-class-out ablation indicates that \texttt{trash can} and \texttt{couch} cause the most significant frequency-normalised decrease in PR-AUC, whereas the most frequent classes, \texttt{computer} and \texttt{chair}, exert only a minor effect.
This pattern is consistent with contrastive sentence encoders that down-weight ubiquitous tokens, such as “the”~\cite{kurita2023contrastivelearningbasedsentenceencoders}.

\subsection{Post-hoc Attribution}
\begin{figure}[bp]
\centering
    \includegraphics[scale=0.5, trim=0 0 0 30, clip]{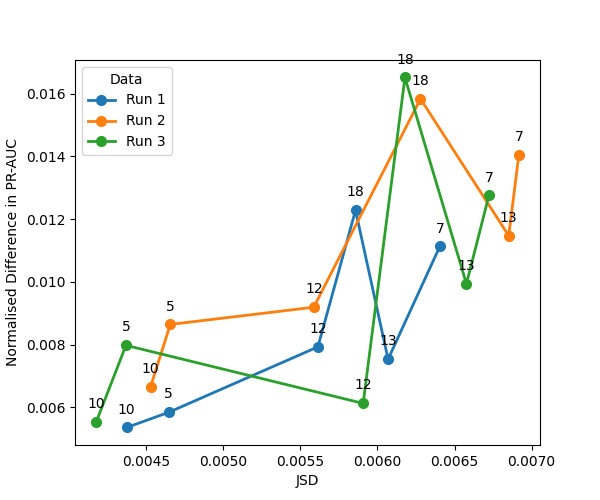} 
    \vspace{-10pt}
 \caption{Normalised change in PR-AUC resulting from removal of each semantic class from the scene graph, measured against the change in node importance distribution obtained from attention weights. The positive correlation confirms that attention can serve as an effective attribution method.}
    \label{fig:attnjsd}

\hfill

\end{figure}

\Cref{tab:all-rankings} summarises class-level importance from all explainers, compared using the previously defined JSD to assess whether frequent classes receive consistently lower attributions. 
The methods broadly agree on the classes most (\texttt{couch}, \texttt{plant}, \texttt{trash can}) and least (\texttt{chair}, \texttt{computer}) useful for localisation.
Most methods rank \texttt{painting} relatively low, with Shapley Value Sampling being the primary exception. \Cref{fig:attnjsd} verifies that Attention Weights correlate with perturbation-induced changes in localisation performance, indicating their effectiveness as feature-importance scores, described in \cref{sec:explainlit}.
Attention shows a clear correlation \cite{fan2021gcnseattentionexplainabilitynode,pana2} and thus is adopted as an explanatory signal.

\begin{table}[!t]
    \centering
    \caption{Semantic-class rankings per method and run. Entries are listed in descending order of importance.}
    \label{tab:all-rankings}
    \setlength{\tabcolsep}{3pt}
    \renewcommand{\arraystretch}{1.05}
    \begin{tabular}{@{}lllllllll@{}}
        \toprule
        \textbf{Method} & \textbf{Run} & \textbf{1st} & \textbf{2nd} & \textbf{3rd} & \textbf{4th} & \textbf{5th} & \textbf{6th} \\
        \midrule
        \multirow{3}{*}{Single-Class Ablation} 
            & 1 & TC & CO & PA & PL & CH & CP \\
            & 2 & TC & CO & PL & PA & CH & CO \\
            & 3 & TC & CO & PL & CH & PA & CP \\
        \midrule
        \multirow{3}{*}{Saliency}
            & 1 & PL & TC & CH & CP & CO & PA \\
            & 2 & PL & CH & CP & TC & CO & PA \\
            & 3 & PL & TC & CO & CH & CP & PA \\
        \addlinespace
        \multirow{3}{*}{Integrated Gradients}
            & 1 & TC & PL & CH & CP & CO & PA \\
            & 2 & TC & PL & CH & CO & CP & PA \\
            & 3 & TC & PL & CH & CO & CP & PA \\
        \addlinespace
        \multirow{3}{*}{Shapley Value Sampling}
            & 1 & CO & PA & TC & PL & CP & CH \\
            & 2 & CO & PA & TC & CP & CH & PL \\
            & 3 & CP & CO & CH & TC & PA & PL \\
        \addlinespace
        \multirow{3}{*}{Attention}
            & 1 & CO & PL & TC & PA & CH & CP \\
            & 2 & CO & PL & TC & PA & CH & CP \\
            & 3 & CO & PL & TC & PA & CH & CP \\
        \bottomrule
    \end{tabular}

\end{table}

Agreement is strongest at the extremes and is stable across runs for Integrated Gradients and Attention Weights. Saliency and Shapley exhibit greater run-to-run variability under environmental perturbations, yet their highest- and lowest-ranked classes typically align with the consensus. To identify the most reliable explainer, each method is further evaluated using fidelity metrics.

\subsection{Fidelity}
Overall, the methods agree at the extremes but diverge in mid-rankings. To identify the most reliable explainer, each method is further evaluated using fidelity metrics.

\begin{figure*}[t]
  \centering
  \begin{subfigure}[t]{0.32\textwidth}
  \includegraphics[width=\textwidth]{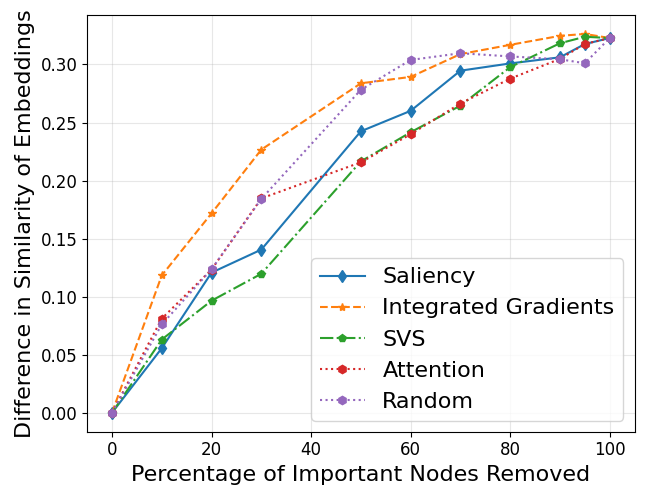}
  \caption{\label{fig:fidelity+}}
  \end{subfigure}
  \begin{subfigure}[t]{0.32\textwidth}
  \includegraphics[width=\textwidth]{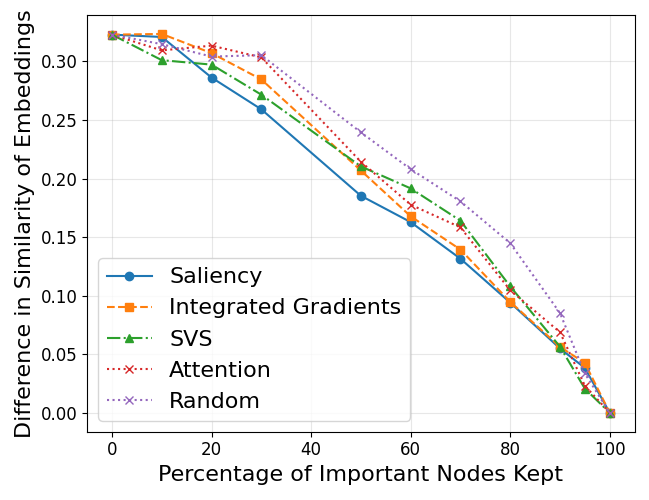}
  \caption{\label{fig:fidelity-}}
  \end{subfigure}
  \begin{subfigure}[t]{0.32\textwidth}
  \includegraphics[width=\textwidth]{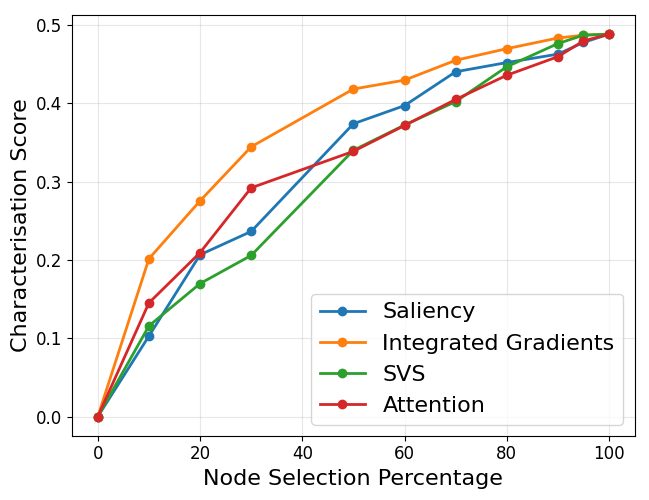}
  \caption{\label{fig:chargarph}}
  \end{subfigure}
  \caption{Fidelity+ (a, necessity of removed nodes), fidelity- (b, sufficiency of kept nodes) and combined characterisation score (c) curves on \textsc{Run~1}. Generally, Integrated Gradients performs best overall, as indicated by its higher characterisation score, where Attention also exhibits a high initial slope, suggesting strong performance in identifying the most critical nodes.}
  \label{fig:fidelity}
  \vspace{-15pt}
\end{figure*}

%

\Cref{fig:fidelity-,fig:fidelity+} plot \(\Delta^-(\rho)\) and \(\Delta^+(\rho)\) averaged over all place–query pairs in run 1, and \cref{fig:chargarph} presents the characterisation scores across different node selection levels (\(\rho\)) for each explainability method. Integrated Gradients consistently achieves the highest characterisation scores, indicating that its top-ranked nodes are both indispensable -- i.e. their removal leads to a significant drop in embedding similarity -- and sufficient, in that only retaining them preserves most of the original similarity.
Attention Weights achieve the second-best scores at lower selection levels, indicating that they are particularly effective in identifying the most important nodes early on.
However, their performance diminishes at higher selection levels.

These results demonstrate that Integrated Gradients and Attention Weights are the most effective explainers for our purposes, especially as their rankings align most closely with the class ablation.
Attention Weights, in particular, offer the added benefits of clear interpretability and minimal computational overhead -- since they are part of the localisation network itself -- making them the preferred method when inference speed or resource constraints are paramount.

\subsection{Relationship with class frequency}

Semantic class rankings from Integrated Gradients, attention and class-ablation ranking (\cref{tab:all-rankings}) can be compared with raw object frequencies (\cref{tab:uhumans2_objects}). Integrated Gradients assigns the highest importance to \texttt{trash can} and \texttt{plant}, and the lowest to \texttt{computer} and \texttt{painting}. Attention, by contrast, ranks \texttt{couch} and \texttt{plant} highest, and \texttt{chair} and \texttt{computer} lowest. The class ablation likewise places \texttt{trash can}, \texttt{couch}, and \texttt{plant} at the top, with \texttt{chair} and \texttt{computer} at the bottom.

\Cref{tab:uhumans2_objects} shows that \texttt{computer} and \texttt{chair} are the most frequent objects, whereas \texttt{painting}, \texttt{couch}, and \texttt{plant} are less common. The results thus indicate an inverse relationship between class frequency and assigned importance, analogous to TF–IDF weighting, whereby high-frequency classes are downweighted and low-frequency classes are upweighted.
Notable exceptions persist: \texttt{trash can} remains highly important despite moderate frequency, and Integrated Gradients assigns relatively low weights to \texttt{painting} instances. Although random perturbations were introduced to \texttt{trash can} and \texttt{chair} during training to model potentially-moveable landmarks, \texttt{trash can} still receives high importance scores.

Overall, object frequency is a \textit{strong but not exclusive} importance indicator, demonstrating that even in a simple localisation model, such as the one we presented, spatial layout and relational context also contribute to the observed importance profiles.

\section{Conclusion, Limitations, and Future Work}
We presented a lightweight, interpretable pipeline for coarse place registration on semantic scene graphs.
By embedding an unperturbed “map” graph and its perturbed “query” into a shared latent space, the model aligns matching place–query pairs. 
Notably, the study places explainability at the centre of semantics-based localisation. A comprehensive analysis, combining fidelity-based evaluations and semantic-class perturbations, identifies Integrated Gradients and Attention Weights as the most faithful explanatory signals for this task.
Class-importance patterns reveal a TF–IDF–like bias, where common objects are often down-weighted.
The high weight of \texttt{trash can} indicates that contextual relations, not just frequency, influence performance.

The results come with certain limitations: the semantics-only setting omits geometry, which enables a focused attribution analysis but caps absolute retrieval performance, and the evaluation is confined to a single synthetic domain (uHumans2 \textsc{office}). 
Future work will integrate lightweight geometric cues, explore more complex model architectures, and evaluate our approach across diverse real-world datasets and semantic taxonomies to stress-test the stability and utility of explanations in deployed localisation systems.

\bibliographystyle{IEEEtran}
\bibliography{mybib.bib}

\end{document}